\documentclass{new_tlp}
\usepackage{mathptmx}
\usepackage[mathscr]{euscript}
\usepackage{comment}
\usepackage{graphicx}
\usepackage{color}

\newcommand\bcmdtab{\noindent\bgroup\tabcolsep=0pt%
  \begin{tabular}{@{}p{10pc}@{}p{20pc}@{}}}
\newcommand\ecmdtab{\end{tabular}\egroup}

     {\begin{list}{}%
             {\setlength{\leftmargin}{#1}}%
             \item[]%
     }
     {\end{list}}

\def\causes{\; {\bf causes} \;}
\def\lif{\; {\bf if} \;}

\def\impossible{{\bf impossible} \;}

\newcommand{\alm}{$\mathscr{ALM}$}
\newcommand{\corealm}{\textsc{CoreALMlib}}
\newcommand{\clib}{\textsc{CLib}}
\newcommand{\aclib}{$a$\textsc{CLib}}

  \title[ \corealm: An \alm\ Library Translated from the \textsc{Component Library}]
        {\corealm: An \alm\  Library Translated from the \textsc{Component Library}}

  \author[D. Inclezan]
         {DANIELA INCLEZAN\\
         Miami University, Oxford OH, 45056, USA\\
         \email{inclezd@miamioh.edu}}

\jdate{March 2003}
\pubyear{2003}
\pagerange{\pageref{firstpage}--\pageref{lastpage}}
\doi{S1471068401001193}

\begin{document}
\nocite{*}

\label{firstpage}

\maketitle

  \begin{abstract}
    This paper presents \corealm, an \alm\  library of commonsense knowledge about dynamic domains.
		The library was obtained by translating part of the \textsc{Component Library} (\clib) 
		into the modular action language \alm. \clib\ consists of general reusable and composable 
		commonsense concepts, selected based on a thorough study of ontological and lexical resources.
		Our translation targets \clib\ {\em states} (i.e., {\em fluents}) and {\em actions}. 
		The resulting \alm\ library contains the descriptions of 123 action classes grouped into 
		43 reusable modules that are organized into a hierarchy. It is made available online and of interest
		to researchers in the action language, answer-set programming, and natural language understanding communities.
		We believe that our translation has two main advantages over its \clib\ counterpart:
		(i) it specifies axioms about actions in a more elaboration tolerant and readable 
		way, and (ii) it can be seamlessly integrated with ASP reasoning algorithms (e.g., for planning and postdiction). 
		In contrast, axioms are described in \clib\ using STRIPS-like operators, and \clib's inference
		engine cannot handle planning nor postdiction. 
		Under consideration for publication in TPLP.
  \end{abstract}

  \begin{keywords}
    action language, commonsense library, knowledge reuse, dynamic domains
  \end{keywords}


\section{Introduction}
\label{intro}

In the field of knowledge representation about dynamic domains,
an important advancement in the last decade 
was the design of so-called {\em modular} action languages,
which focus on the reuse of knowledge and creation of libraries
in general, and the representation of actions in terms of other actions in particular.
\alm\ (Action Language with Modules) \cite{ig16} was introduced to address these issues. 
The capabilities of the language were tested in the context of Project Halo 
\cite{projecthalo10}: specialized knowledge about a biological process was encoded in
\alm\ and then used to answer non-trivial questions \cite{ig11}.

After defining our language, our next goal was to develop an \alm\ 
library of commonsense knowledge that would facilitate the description
of large dynamic systems through the reuse of its components.
We started by creating a small library of motion. 
We were satisfied with the result, and were able to use our representation
in solving reasoning tasks by combining system descriptions of \alm\ 
with reasoning algorithms written in Answer Set Prolog (ASP) \cite{gl88,gl91}.
However, we were not sure what criteria to use for the
selection of further concepts, general enough to deserve inclusion in our 
library.  

An answer to this question was provided by our collaboration on project AURA
(Automated User-centered Reasoning and Acquisition System)
\cite{cjmpps07,ccbchjpsty07,ccmpst09}.
AURA is a knowledge acquisition system that allows domain experts to enter
knowledge and questions related to different disciplines, with minimal
support from knowledge engineers.
Our task within the project was to revise a section of AURA's core knowledge base, 
called {\sc Component Library} (\clib), from the point of view of 
its soundness and completeness with respect to the goals of AURA \cite{cdi14}.
\clib \ \cite{bpc01} is a vast library of general, reusable knowledge components
with goals similar to ours.
It was extensively tested in AURA along the years and,
more importantly for us, its concepts were selected using a well-founded methodology,
based on lexical and ontological resources.
We benefited from these key features of \clib \ by porting part of the library, 
in its revised form resulting from our analysis, into \alm. 
The resulting \alm\ library combines the advantages of \clib \ with
established knowledge representation methodologies
developed in the {\em action language} community:
concise and elaboration tolerant representations of action effects and preconditions,
and an easy coupling with reasoning algorithms encoded in ASP \cite{gl98}.
In the future, the experience of creating a \clib-inspired library 
will allow us to further explore how \alm\ libraries should be 
structured, queried, used, and made available to the public.

\paragraph{{\bf Related Work.}}

A previous version of this work was presented in an earlier paper \cite{di15}
whose goal was to introduce \alm\ and explain
how enforcing the thought pattern of our language on a translation from \clib\
results in much more concise representations. Instead, in the current paper
we describe in more detail a translation that is closer to \clib\ and can potentially be automated.
We also make the library available online and discuss how it can be used.

In addition to \clib, several linguistic resources, including VerbNet \cite{KipperPhd05}, 
WordNet \cite{mil90} and FrameNet \cite{framenet},
contain information about verbs, sometimes organized into an inheritance hierarchy or
into sets of synonyms. Verbs may sometimes be accompanied by axioms, but these are
expressed in informal terms, and therefore cannot be readily used to create an \alm\ library
of commonsense knowledge.
A modular action language with similar goals to \alm's is MAD \cite{lr06,el06}.
There is a collection of MAD libraries \cite{thesiser08}, but its fifteen library modules
only describe eight action classes (compared to 146 in \clib), 
and are less expressive as a result.


\section{Modular Action Language \alm}
\label{alm}

A dynamic system is represented in \alm\ 
by a {\em system description} that consists of two parts:
a {\em general theory} (i.e., a collection of modules organized into
a hierarchy) and a {\em structure} (i.e., an interpretation of 
some of the symbols in the theory). 
A {\em module} is a collection
of declarations of sorts (i.e., classes) and functions together with a set of axioms.
The purpose of a module is to allow the organization of knowledge into smaller 
reusable pieces of code. 
Modules serve a similar role to that of procedures in procedural languages
and can be organized into a hierarchy (a DAG).
If a module $M_1$ contains the statement 
``{\bf depends on} $M$",
then the declarations and axioms of module $M$ are implicitly part of $M_1$.

We briefly illustrate the syntax of \alm\ via some examples.
For a formal description of the language, we direct the reader
to the paper that introduces \alm\ \cite{ig16}.
Boldface symbols denote keywords of the language;
identifiers starting with a lowercase letter denote constant symbols;
and identifiers starting with an uppercase letter denote variables.

Sorts are organized into a hierarchy with root ${\bf universe}$ and
pre-defined sorts ${\bf actions}$ and ${\bf booleans}$. The sort hierarchy is specified  
via the specialization construct ``::''. For instance, we say that
$points$ and $things$ are subsorts of ${\bf universe}$ and $agents$ is a subsort of
$things$ by: 

\noindent
$
\begin{array}{l}
\ \ \ \ \ \ \ points, things\ :: \ {\bf universe}\\
\ \ \ \ \ \ \ agents \ :: \ things
\end{array}
$

\noindent
We use the same construct to define action classes, if needed as special cases
of other action classes. Declarations of action classes
include the specification of {\em attributes}, which are intrinsic properties.
For instance, the statements:

\noindent
$
\begin{array}{l}
\ \ \ \ \ \ \ move \ :: \ {\bf actions}\\
\ \ \ \ \ \ \ \ \ \ \ {\bf attributes}\\
\ \ \ \ \ \ \ \ \ \ \ \ \ \ \ actor : agents\\
\ \ \ \ \ \ \ \ \ \ \ \ \ \ \ origin, dest : points
\end{array}
$

\noindent
define $move$ as having three attributes,
$actor$, $origin$, and $dest$ -- (possibly partial) functions mapping elements of $move$
into elements of $agents$, $points$, and $points$ respectively.

Properties of objects of a dynamic system are represented using functions. 
Functions are partitioned in \alm\ into {\em fluents}
(those that can be changed by actions) and {\em statics} (those that cannot); each of these
are further divided into {\em basic} and {\em defined}, where defined functions
can be viewed just as means to facilitate
knowledge encoding. Basic fluents are subject to the law of inertia.
In our example, the location of $things$ is encoded by the

$
\begin{array}{l}
\ \ \ \ {\bf fluent}\ {\bf basic} \ loc\_in : things \rightarrow points.
\end{array}
$

Axioms that can appear in a module are of four types -- 
dynamic causal laws, state constraints, definitions for defined functions, 
and executability conditions. 
The axiom:

$
\begin{array}{l}
\ \ \ \ occurs(X) \ \causes \ loc\_in(A) = D \ \lif \ instance(X, move), \ actor(X) = A,\ dest(X) = D.
\end{array}
$

\noindent
is a dynamic causal law saying that the actor of a $move$ action will
be located at the destination after the execution of the action.

The second part of a system description is its {\em structure}, which represents 
information about a specific domain: instances of sorts (including actions)
and values of statics.
For example, a domain that is about John and Bob, and their
movements between two points, $a$ and $b$, may be described as follows:

\noindent
$
\begin{array}{l}
\ \ \ \ \ \ \ john, bob \ {\bf in} \ agents\\
\ \ \ \ \ \ \ a, b \ {\bf in} \ points
\end{array}
$

\noindent
$
\begin{array}{l}
\ \ \ \ \ \ \ go(X, Y) \ {\bf in} \ move\\
\ \ \ \ \ \ \ \ \ \ actor = X\\
\ \ \ \ \ \ \ \ \ \ dest = Y
\end{array}
$

\noindent
Action $go(X, Y)$ is an instance schema that stands for all actions
of this form obtained by replacing $X$ and $Y$ with instances of $agents$ and $points$, respectively. 
One such action is $go(john, a)$ for which
attributes have the values $actor(go(john, a)) = john$ and 
$dest(go(john, a)) = a$.

The semantics of \alm\ is given
by defining the states and transitions of the transition diagram 
defined by a system description. 
For that purpose, we encode statements of the system
description into a logic program of ASP\{f\} \cite{bal13}, an
extension of ASP by non-Herbrand functions.
The states and transitions of the corresponding transition diagram 
are determined by parts of the answer sets of this logic program. 
As an example, the dynamic causal law about actions of the type $move$ shown above 
is encoded as:

\noindent
$
\begin{array}{l}
\ \ \ \ \ \ \ loc\_in(A, I+1) = D \ \leftarrow \ instance(X, move), \ occurs(X, I), \ 
                    actor(X) = A,\ dest(X) = D
										
\end{array}
$

\noindent
followed by replacing variables (other than $I$) by constants of the appropriate sorts. 
The structure is encoded using statements like:

\noindent
$
\begin{array}{l}
\ \ \ \ \ \ \ is\_a(john, agents).
\end{array}
$

\noindent
$
\begin{array}{l}
\ \ \ \ \ \ \ is\_a(go(john, a), move).\ \ \ \ actor(go(john,a)) = john.\ \ \ \ dest(go(john,a)) = a.\ \ \ \ (etc.)\\
\end{array}
$


\section{\clib \ and KM}
\label{clib}

\clib \ \cite{bpc01} is a library of general,
reusable, composable, and interrelated components of knowledge.
Notions included in \clib \ were selected using a solid methodology
relying on linguistic and ontological resources such as WordNet (wordnet.princeton.edu),
FrameNet (framenet.icsi.berkeley.edu),
VerbNet (verbs.colorado.edu/verb-index),
a thesaurus and an English dictionary, as well as various 
ontologies from the semantic web community.
\clib \ was built with three main design criteria in mind:
(1) {\em coverage}: \clib \ should contain enough components to allow representing a variety of knowledge;
(2) {\em access}: components should meet users' intuition and be easy to find; and
(3) {\em semantics}: components should be enriched with non-trivial axioms.
These are criteria that we want for a core \alm\ library as well.

The \clib\ library is written in the knowledge representation
language of Knowledge Machine (KM) \cite{cp04}.
KM is a frame-based language with first-order logic (FOL) semantics.
The language uses its own syntax, which is mainly syntactic sugar for 
FOL \cite{cp04}.
KM distinguishes between two basic concepts: {\em class} and {\em instance}. 

In \clib, there are three main classes -- {\tt Entity}, {\tt Event}, and {\tt Role} --
and the additional built-in class {\tt Slot}. All are subclasses of the root class {\tt Thing}.
Events are divided into {\tt Action}s, {\tt Activit}ies, and {\tt State}s  
(which should not be confused with states of the transition diagram).
An example of a \clib\ action is {\tt Obstruct}, whose declaration can be seen in Figure \ref{fig1}(a).
It starts with the specification of properties of the class itself:
it is a subclass of the action class {\tt Make-Inaccessible} and it has an associated 
set of WordNet 2.0 synonyms that includes the verbs ``obstruct" and ``jam." 
In the second part of the declaration, properties of {\em instances} of the action class are specified.
It is asserted that each instance of {\tt Obstruct} has an {\tt object} that 
must be an instance of {\tt Tangible-Entity}. In practice, this results in the automatic
creation of a Skolem constant named {\tt \_Tangible-Entity$N$} to denote 
the object of an {\tt Obstruct} instance, where $N$ is a number.
Next, {\tt resulting-state} and {\tt add-list} together specify 
the effect of the execution of an {\tt Obstruct} action: the object will be obstructed 
(by the agent, if it was defined).
The default {\tt preparatory-event} says that, if the {\tt Obstruct} action has an agent,
then the agent should be where the object is, in order for the action execution
to be possible. Note that failure to meet default axioms only generates warnings and 
does not prevent actions from actually being executed. 
The declaration of {\tt Obstruct} also contains the description of a test case, information 
for generating English text, and a WordNet 1.6 synset, which were not included in the figure.

\begin{figure}[!htbp]
\small
(a) The Declaration of the Action Class {\tt Obstruct}
\begin{verbatim}
(Obstruct has
  (superclasses		(Make-Inaccessible))
  (wn20-synset ((:set
    (:triple "obstruct" 2 "v") (:triple "block" 10 "v") (:triple "close_up" 1 "v")
    (:triple "jam" 7 "v")      (:triple "impede" 2 "v") (:triple "occlude" 1 "v")))))
(every Obstruct has
  (object ((a Tangible-Entity)))
  (resulting-state ((a Be-Obstructed)))
  (add-list (
    (:triple (the resulting-state of Self) object (the object of Self))
    (if (has-value (the agent of Self))
     then (forall (the agent of Self)
			      (:triple It agent-of (the resulting-state of Self))))))
  (preparatory-event (((:default
    (if (has-value (the agent of Self))
     then (a Move with 
		        (object 	   ((the agent of Self)))
		        (destination ((a Spatial-Entity with (is-at ((the object of Self)))))))))))))
\end{verbatim}
\smallskip
\rule{\textwidth}{0.4pt}
(b) The Declaration of the State Class {\tt Be-Obstructed}
\begin{verbatim}
(Be-Obstructed has
  (superclasses   (Be-Inaccessible))
  (wn20-synset ((:set (:triple "obstructed" 1 "a")))))
(every Be-Obstructed has
  (object ((a Entity))))
\end{verbatim}
\smallskip
\rule{\textwidth}{0.4pt}
\begin{tabular}{l|l}
\begin{minipage}{.48\textwidth}
\smallskip
(c) The Definition of the Slot Instance {\tt is-at}
\begin{verbatim}
(is-at has
  (instance-of (Spatial-Relation))
  (domain (Spatial-Entity))
  (range (Spatial-Entity))
  (cardinality (N-to-N))
  (fluent-status (*Inertial-Fluent)))
\end{verbatim}
\end{minipage}
&
\begin{minipage}{.48\textwidth}
\smallskip
(d) The Definition of the Slot Instance {\tt object}
\begin{verbatim}
(object has
  (instance-of (Participant-Relation))
  (domain (Event))
  (range (Entity))
  (cardinality (N-to-N))
  (fluent-status (*Inertial-Fluent)))
\end{verbatim}
\end{minipage}
\end{tabular}
\normalsize
\caption{Examples of Declarations/ Definitions of \clib\ Components}
\label{fig1}
\end{figure}

Expressions like {\tt superclasses} and {\tt object} in the declaration of {\tt Obstruct} are {\em slots}
(i.e. instances of the class {\tt Slot}): 
they denote binary relations that hold between the class or instances of the class 
and the value denoted by the expressions that follow. 
Two of the slots, {\tt superclasses} and {\tt add-list},
are built-ins of KM, while the other ones are specific to \clib.
The definitions of slots {\tt is-at} and {\tt object} 
can be seen in Figure \ref{fig1}(c) and (d) respectively. 

States of \clib\ correspond to fluents in action language terminology, but they are
represented by classes, which means that they can be organized into an inheritance hierarchy. 
An example of a \clib\ state is {\tt Be-Obstructed} from Figure \ref{fig1}(b)
that is a subclass of the state {\tt Be-Inaccessible} and corresponds to the first sense 
of the adjective ``obstructed'' in WordNet 2.0.
Every instance of {\tt Be-Obstructed} is asserted to have an {\tt object} 
that must be an {\tt Entity}.

The \clib \ library was integrated in two systems developed 
at SRI International, SHAKEN and its successor AURA \cite{cjmpps07,ccbchjpsty07}, 
and was extensively tested as a result.
In AURA, domain experts were able to encode new knowledge by building upon the general concepts
of {\sc CLib} in a speedy manner and with no or minimal involvement from knowledge engineers.
Their encodings were used to answer questions from an Advanced Placement 
test suite. At least 70\% of these questions were accurately answered by the AURA system 
in all three domains of interest \cite{projecthalo10}. 
These results seem to demonstrate that {\sc CLib} is a valuable library of general concepts.

\section{A (Partial) Translation from KM into \alm}
\label{translation}

In this work, we only intend to translate part of \clib\ into \alm\ and
refer to this portion as \aclib. 
Specifically, \aclib\ focuses on actions and states, but ignores roles, 
activities, actions with subevents or duration, and text generation information
associated with \clib\ components,
which either deserve more study (roles, actions with duration) or 
exceed the scope of this project.
Thus, \aclib\ contains all but 23 of the 146 actions of \clib\ and 
all but four of its 33 states.
The \aclib\ concepts were manually translated into \alm.
The translation presented here covers the fragment of KM used in \aclib.
To give a simple example, in KM, one may assert that a slot of a specific instance maps into 
a given number of other instances by using the expression ``{\tt exactly $n$}'';
however in all occurrences of this expression in \aclib, $n$ only ranges over \{0, 1, 2\}.

Two main differences between KM and \alm\ posed problems to the translation:
(1) the unique name assumption is part of the semantics of \alm, but
not of KM and 
(2) FOL quantifiers do not exist in \alm, but they do in KM.
The existential quantifier in KM, ``{\tt (a <c>)}", 
is constructive and introduces a Skolem constant,
as explained in Section \ref{clib}. KM implements a complex unification algorithm
for automatically introduced Skolem constants.
In \alm, the user of the library introduces and names instances when specifying the structure
of a system description. We translated the existential quantifier
using constraints to produce a similar effect to KM.

An additional difficulty stems from the fact that \clib\ states (i.e., fluents) 
are represented as classes and thus can be organized 
into a hierarchy and have an arbitrary number of parameters. 
Instead, fluents of \alm\ are represented as functions and have a fixed arity.
To address these problems, we generally translated one \clib\ state using
two \alm\ fluents with different arities and added state constraints
to encode the inheritance relation between \clib\ states.

In what follows, we briefly describe the translation from KM into \alm.
We start with definitions of slots, then move on to declarations of states and actions.
We ignore the difference between the capitalization conventions of KM versus \alm\ 
for conciseness and clarity.
In KM code, we put placeholders for constants inside chevrons ({\tt <>}); 
in \alm\ code, we omit the chevrons.

\subsection{Slots}

Consider a slot {\tt s} defined as follows:
\small
\begin{verbatim}
(<s> has (instance-of (<c>))       
         (domain (<c1>))               
         (range (<c2>))
         (cardinality (<card>))    
         (fluent-status (<status>)))
\end{verbatim}
\normalsize
If {\tt card} is {\tt N-to-N}, then it is translated as the \alm\ function 

{\tt s} $:$ {\tt c1} $\times$ {\tt c2} $\rightarrow {\bf booleans}$

\noindent
and if {\tt card} is {\tt N-to-1} as 

{\tt s} $:$ {\tt c1} $\rightarrow$ {\tt c2}.

\noindent
There are no slots with cardinality {\tt 1-to-N} in \aclib.
If {\tt status} is {\tt *Non-Fluent}, then {\tt s} is a basic static in \alm.
Otherwise, it is a basic (i.e., inertial) fluent, unless:

(1) {\tt c} is {\tt Participant-Relation} or

(2) {\tt c} is {\tt Spatial-Relation} and {\tt c1} is {\tt Event}.

\noindent
As an example, the slot {\tt is-at} from Figure \ref{fig1}(c) is encoded as the
basic fluent 

$is\_at : spatial\_entity \times spatial\_entity \rightarrow {\bf booleans}$.

\noindent
If conditions (1) or (2) above are satisfied, then {\tt s} 
describes an intrinsic property of an event (and thus of an action too) and 
is translated as an attribute of the pre-defined \alm\ class ${\bf actions}$.

\subsection{States}

\clib\ states encode fluents as classes. They do not 
have a fixed arity, rather their parameters are represented via slots that are instances
of the subclass {\tt Participant-Relation} of {\tt Slot}  (e.g., {\tt object}, {\tt instrument}). 
Certain participant relations are {\em required} for a state:
if the declaration of a state {\tt c} says that 

{\tt (every <c> has (<r> ((a <c1>))))}

\noindent
then {\tt r} is required for {\tt c}.
We analyzed participant relations w.r.t. states and
determined that (i) {\tt object} was required in all states; 
(ii) a few states had a secondary required participant relation 
(e.g., {\tt base} in {\tt Be-Confined} referring to the place where the object is confined); and 
(iii) a few other participant relations were only sometimes associated with some of the states 
(e.g., {\tt instrument} in {\tt Be-Blocked} denoting the thing with which the object 
is blocked). 
As a result, if a state class {\tt Be-<f>} only had one required participant relation, {\tt object},
we translated it by creating a basic fluent

$is\_${\tt f} $:$ {\tt c1} $\rightarrow {\bf booleans}$

\noindent
where {\tt c1} is the most specific class membership information for the
participant relation {\tt object} with respect to the state {\tt Be-<f>}. For instance,
for {\tt Be-Obstructed}, we created the basic fluent

$is\_obstructed : spatial\_entity \rightarrow {\bf booleans}$

\noindent
If the state had a second (required/ associated) participant relation {\tt r} with range {\tt c2}, 
we created (instead/ in addition, respectively)  the basic fluent

{\tt f}$\_${\tt prep} $:$ {\tt c1} $\times$ {\tt c2} $\rightarrow {\bf booleans}$

\noindent
where {\tt prep} is the preposition that is normally associated in English with the participant relation 
{\tt r} (e.g., for {\tt instrument}, it would be the preposition ``with").
As an example, the {\tt Be-Blocked} state would be translated by two fluents:

$is\_blocked : spatial\_entity \rightarrow {\bf booleans}$

$blocked\_with : spatial\_entity \times entity \rightarrow {\bf booleans}$

Whenever we translated a \clib\ state by introducing two \alm\ fluents with different
arities, we connected the two via axioms of the style:

$
\begin{array}{lll}
is\_blocked(O) & \lif & blocked\_with(O, I).\\
\neg blocked\_with(O, I) & \lif & \neg is\_blocked(O).				
\end{array}		
$

We expressed the subclass relationship between \clib\ states by introducing
state constraints. For example, we said that {\tt Be-Obstructed} is a subclass of {\tt Be-Inaccessible}
via the axiom:

$
\begin{array}{lll}
\neg is\_accessible(O) & \lif & is\_obstructed(O). 
\end{array}		
$

\subsection{Actions}

We replaced the \clib\ class {\tt Action} with the predefined \alm\ class \textbf{actions}.
We translated the information about the superclasses of an action class of \aclib
\begin{verbatim}
  (<c> has (superclasses (<s1> ... <sn>))
\end{verbatim}

\noindent
using the specialization construct (i.e., ``::") of \alm:

${\tt c} :: {\tt s1},\ \dots,\ {\tt sn}$.

\noindent
The information about WordNet 2.0 synsets was integrated in the online tool (see Section \ref{lib}).

The description of properties of every instance of the class
can be divided into two parts that we address in separate subsections below:
(1) {\em attribute declarations} -- the description
of the values of participant or spatial relations like {\tt object}, {\tt agent}, {\tt origin},
{\tt destination}, etc. that describe intrinsic properties of the action; and 
(2) {\em axioms} -- the specification of the values of slots {\tt resulting-state}, 
{\tt defeats}, {\tt add-list}, {\tt del-list}, {\tt pcs-list}, etc. that describe
the effects and preconditions for the execution of the action.

\subsubsection{Attribute Declarations}
\label{attributes}

In what follows, by {\tt <attr>} we denote a slot that stands for a participant or spatial relation
describing an inherent property of instances of the action class and refer to it as an
{\em attribute}. Consider the following
part of the declaration of action class {\tt c}:
\begin{verbatim}
  (every <c> has (<attr> ((a <c1>))))
\end{verbatim}
In KM, this generates a Skolem constant for each instance of {\tt c},
which cannot be done in \alm. We encode the statement
via an attribute declaration (when needed) and several axioms, one of them
requiring the introduction of a defined fluent, as explained below.
If ${\tt attr}$ is not an attribute of any superclass of ${\tt c}$, 
we add to the declaration of ${\tt c}$ the attribute:

$
\begin{array}{l}
{\tt attr} : {\tt c1} \rightarrow \textbf{booleans}
\end{array}
$

\noindent
If ${\tt attr}$ is an attribute of a superclass ${\tt s}$ of ${\tt c}$, 
translated in \alm\ as
${\tt attr} : {\tt c2} \rightarrow \textbf{booleans}$
such that ${\tt c2}$ is different from ${\tt c1}$ (possibly a superclass of ${\tt c1}$),
we add the constraint

$false \ \ \lif \ instance(X, {\tt c}),\ \ \ {\tt attr}(X, A),\ \ \neg instance(A, {\tt c1}).$

\noindent
In both cases, we add a defined static 

$defined\_{\tt attr} : {\tt c} \rightarrow \textbf{booleans}$ 

\noindent
that is true when there is a value for the attribute and require it to be true for all instances of {\tt c}:

$defined\_{\tt attr}(X) \ \ \lif \ \ {\tt attr}(X).$

$false \ \ \lif \ \ instance(X, {\tt c}),\ \ \neg defined\_{\tt attr}(X).$

\smallskip
Constraints on the values of attributes in \aclib\ are encoded
using state constraints of \alm, sometimes preceded by the declaration of new
defined statics that are needed to deal with the lack of quantifiers in \alm.
The constraint that {\tt attr} must range over instances of {\tt c1}
\begin{verbatim}
  (every <c> has (<attr> ((must-be-a <c1>))))
\end{verbatim}
is translated as

$false \ \ \lif \ \ instance(X, {\tt c}),\ {\tt attr}(X, A),\ \neg instance(A, {\tt c1}).$

\noindent
Similarly if the keyword {\tt mustnt-be-a} is used instead.
The constraint that {\tt attr} must map into at most one instance of {\tt c1}
for instances of {\tt c} (after unification):
\begin{verbatim}								
  (every <c> has (<attr> ((at-most 1 <c1>))))
\end{verbatim}
is captured by the state constraint

$\neg {\tt attr}(X, A1) \ \ \lif \ \ {\tt attr}(X, A2),\ 
                        A1 \neq A2,\ 
                        instance(X, {\tt c}),\ 
                        instance(A1, {\tt c1}),\ 
                        instance(A2, {\tt c1}).
$																	
				
\noindent																	
Similarly for {\tt at-most 2}, just with a more complex axiom.
If {\tt at-most} is substituted by {\tt at-least}, we have axioms similar to the ones for
{\tt (a <c1>)} in \alm. For {\tt at-least 2}, we would first introduce a defined fluent 
$at\_least\_2\_{\tt attr} : {\tt c} \rightarrow \textbf{booleans}$
that is true whenever there are at least two distinct values of class ${\tt c1}$ for {\tt attr};
then we would add the state constraint

$false \ \ \lif \ \ instance(X, {\tt c}), \ \neg at\_least\_2\_{\tt attr}(X).$

\noindent
The keyword {\tt exactly $n$} in a constraint would be translated by putting together
the translations of {\tt at-least $n$} and {\tt at-most $n$} if {\tt $n$} is 1 or 2. 
If {\tt $n$} is 0, we would add the state constraint

$false \ \ \lif \ \ instance(X, {\tt c}),\ 
       {\tt attr}(X, A),\ 
       instance(X, {\tt c1}).
$

\smallskip
Other constraints require the value of an attribute to be the same, different,
or unifiable (``{\tt \&}") with that of some other attribute or expression. For the constraints 

(a) {\tt (every <c> has (<attr1> ((the <attr2> of Self))))}

(b) {\tt (every <c> has (<attr1> ((excluded-values (the <attr2> of Self))))}

(c) {\tt (every <c> has (<attr1> ((constraint (TheValue \& (the <attr2> of Self))))))}

\noindent
the corresponding state constraints in the \alm\ translation are

(a) ${\tt attr1}(X, V) \ \ \lif \ \ {\tt attr2}(X, V),\ instance(X, {\tt c}).$

(b) $false \ \ \lif \ \ instance(X, {\tt c}),\ {\tt attr1}(X, A),\ {\tt attr2}(X, A).$

(c) In \alm, we must say that the values are the same (via a defined static and several axioms).

\subsubsection{Axioms} 

We present the translation of the most common types of axioms
encountered in \aclib. More complex axioms were translated in a similar way.

\smallskip
\noindent
\textbf{Action Preconditions} were specified using lists of properties that 
must be true ({\tt pcs-list}) or false ({\tt ncs-list}) for the action to be executable.
Consider the axiom:
\small
\begin{verbatim}	
(every <c> has (pcs-list ((forall (the <attr> of Self)
                             (:triple It object-of (a Be-<f>)))))
\end{verbatim}
\normalsize
where {\tt object-of} is the inverse of {\tt object}.
It says that, in order for an action of class {\tt c} to be executable, its attribute {\tt attr}
must be in a {\tt Be-<f>} state. Thus we translate it as:

$
\impossible \ \ occurs(X) \ \ \lif \ \ instance(X, {\tt c}),\ 
            {\tt attr}(X, A),\ 
						\neg is\_{\tt f}(A).
$					

\noindent
Another common type of axiom was the one below:
\small
\begin{verbatim}	
(every <c> has 
  (pcs-list 
    ((forall (the <attr1> of Self)
      (:triple It object-of (a Be-<f> with (<attr2> ((the <attr2> of Self)))))))))
\end{verbatim}				
\normalsize
This says that the action's attribute {\tt <attr1>} must
be in a {\tt Be-<f>} state that has attribute {\tt attr2} mapped into
the action's same attribute. If we denote by 
{\tt prep} the preposition associated with the attribute {\tt attr2}
in the context of state {\tt Be-<f>}, then the \alm\ translation
would look as follows:
				
$\impossible \ \ occurs(X) \ \ \lif \ \ instance(X,$ \small{\tt c}\normalsize),\ 
            \small{\tt attr1}\normalsize$(X, A1)$,\ 
						\small{\tt attr2}\normalsize$(X, A2)$,\
						$\neg$ \small{\tt f}\_{\tt prep}\normalsize$(A1, A2).$					

\noindent
Negative preconditions ({\tt ncs-list}s) are translated in a similar way.


\medskip
\noindent
\textbf{Action Effects} were normally encoded by a {\tt resulting-state} 
referenced in an {\tt add-list} (properties that will hold
after the action execution), or a {\tt defeats} state referenced in a {\tt del-list}
(properties that will not hold).
The simplest form of an {\tt add-list} of \aclib\ can be seen below:
\small
\begin{verbatim}	
(every <c> has 
  (resulting-state ((a Be-<f>)))
  (add-list 
	   ((:triple (the resulting-state of Self) object (the <attr> of Self)))))
\end{verbatim}
\normalsize
saying that as a result of the execution of an action of class {\tt c},
its attribute {\tt attr} will now be the object of a {\tt Be-<f>} state.
Its translation into \alm\ is:

$
occurs(X) \ \ \causes \ \ is\_{\tt f}(A) \ \ \lif \ \ instance(X, {\tt c}),\ 
          {\tt attr}(X, A).
$			

A more complex form, containing an {\tt if} -- {\tt then} expression is:
\small
\begin{verbatim}	
(every <c> has
  (resulting-state ((a Be-<f>)))
  (add-list (
	  (if (has-value (the <attr2> of Self))
     then (:triple (the resulting-state of Self) <attr2> (the <attr2> of Self))
     else (:triple (the resulting-state of Self) object (the <attr1> of Self))))))
\end{verbatim}
\normalsize
In addition to what was mentioned for the previous axiom, this says that,
if there is value for the action's attribute {\tt attr2},
then it will also be the {\tt attr2} of the {\tt Be-<f>} state.
This requires using the binary version of the corresponding fluent in \alm.
Assuming that {\tt prep} is the preposition associated with attribute {\tt attr2} in 
the binary version of {\tt f}, then the \alm\ translation would be:

$
occurs(X) \ \ \causes \ \ {\tt f}\_{\tt prep}(A1, A2) \ \ \lif \ \ instance(X, {\tt c}),\
          {\tt attr1}(X, A1),\ 
          {\tt attr2}(X, A2).
$	

$
occurs(X) \ \ \causes \ \ is\_{\tt f}(A) \ \ \lif \ \ instance(X, {\tt c}),\ 
           {\tt attr1}(X, A),\ 
					 \neg defined\_{\tt attr2}(X).
$	

Finally let us consider an example of a {\tt del-list}:
\small
\begin{verbatim}
(every <c> has
  (defeats ((allof (the object-of of (the <attr> of Self))
			       where ((the classes of It) = Be-<f>))))
  (del-list ((forall (the defeats of Self)
                 (:triple (It) object (the <attr> of Self) )))))
\end{verbatim}
\normalsize
This says that the action's attributes {\tt attr} that are in a {\tt Be-<f>} state
should no longer be in this state after the execution of the action.
We translate it in \alm\ as:

$	
occurs(X) \ \ \causes \ \ \neg is\_{\tt f}(A) \ \ \lif \ \ instance(X, {\tt c}),\ 
          {\tt attr1}(X, A),\ is\_{\tt f}(A).
$


\medskip
\noindent
\textbf{Defeasible Axioms.} Some action classes of \aclib\ also contained 
a set of axioms that were defaults (they did not prevent the action from being executed,
just produced warnings). They were specified as {\tt soft-pcs-list} or {\tt preparatory-event}s.
We translated both in a similar way to {\tt pcs-list}, but marked them as
optional when including their translation in the library, as further discussed 
in Section \ref{lib}. This allows the user to decide whether they apply to a particular 
domain to be represented and should be included in the system description or not.

\subsection{Remarks}
In addition to the direct translation described above, we sometimes added extra axioms
to the \alm\ translation for consistency or in order to account for specifications 
that we felt were missing.
An example of the first is when one action class contained an executability condition
but the action class with an opposite effect did not contain the counterpart
(e.g., {\tt Unblock} and {\tt Block}).
For {\tt Move} we added the restrictions that the origin and destination should have at most one value,
to distinguish it form action classes that would otherwise be more suitable 
(e.g., {\tt Move-Together} or {\tt Move-Apart}).
Additionally, since we focused on discrete actions that occur instantaneously,
we changed the names of a couple of actions to make them sound less as processes
and more like discrete actions (e.g., {\tt Hold} was renamed as $take\_hold$).

Axioms in the \alm\ translation are more succinct and elaboration tolerant
that the STRIPS-like add and delete lists used in \clib. On the other hand,
constraints on attributes are expressed more concisely in KM using  
FOL quantifiers; in \alm\ axioms and extra statics are needed.
This can be easily addressed in \alm\ by adding extra keywords 
and expanding the language with aggregates in the spirit of Gelfond and Zhang \citeyear{gz14}, 
which we plan to do in  the near future.

Formulating and proving a formal result on the soundness of 
our translation is non-trivial because KM is not exactly FOL and thus
its semantics are not completely clear.
This task deserves further investigation that will be the subject of
another paper.
We can say however that the translation is faithful to the intended
meaning of \aclib\ concepts.

\section{The Translated Library \corealm}
\label{lib}

\subsection{Organizing the Library into Modules}

The translation from KM (as used in \aclib) into \alm\ was the first step in 
producing an \alm\ library of core commonsense knowledge. One additional step was needed however,
because the basic concept of \alm, that of a {\em module}, is a higher-level one than the main concepts
of KM ({\em class} and {\em instance}). An \alm\ module is a reusable piece of knowledge
on a specific theme that groups together declarations of classes (including action classes) and functions,
as well as axioms about these. \clib\ has no concept similar to a module.
Determining what modules to create, what functions and action classes of the translation 
to include in each module, and how to organize modules into a dependency hierarchy 
was not a trivial task. 

We started by applying the guidelines for creating modular \alm\ representations provided
by Inclezan and Gelfond \citeyear{ig16}. 
Specifically, we started from the top of the \aclib\ class hierarchy, and
gradually built and tested modules capturing knowledge about actions, 
while reusing previously written modules as much as possible.
This required us to create a root module that contained the translations of classes
{\tt Event} and {\tt Action} from \aclib\ and the general part of the {\tt Entity} hierarchy.
All possible participant and spatial relations
of an action (i.e., {\em attributes} in \alm\ terminology) were also included in this module 
called $entity\_event\_and\_action$. 
We identified fifteen major themes that allowed us to group the specific action classes 
into modules. 
For this purpose, we analyzed: 
(i) the hierarchies of state and action classes;
(ii) the fluents relevant to each action class; and
(iii) the list of neighboring concepts listed in the \clib\ user interface,
which sometimes included an opposite action (e.g., {\tt Unobstruct} for {\tt Obstruct}).
We placed in the same module opposite actions, actions that affected or had preconditions 
described in terms of the same \clib\ state, and sometimes subclasses of an action class.
We strove to balance the size of a module with the depth of the (part of the) 
action class hierarchy that it captured and also considered the
resulting depth of the module dependency hierarchy. Because of this, we sometimes
placed subclasses of an action class in a separate module.
Optional axioms (i.e., axioms resulting from the translation of default statements of \aclib) 
and any needed declarations
were placed in a separate leaf module that depended on the module containing the declaration of 
the action class.
The resulting library, \corealm, consists of 43 modules. 
It encodes information about 123 action classes.
Each module contains the description of one to six action classes. 
The depth of the module hierarchy is three (four if counting
modules containing optional axioms), which we believe to be manageable.

\subsection{Online Tool}

\begin{figure}[!htbp]
\centering
\setlength\fboxsep{0pt}
\setlength\fboxrule{0.5pt}
\fbox{\includegraphics[width=1\textwidth]
{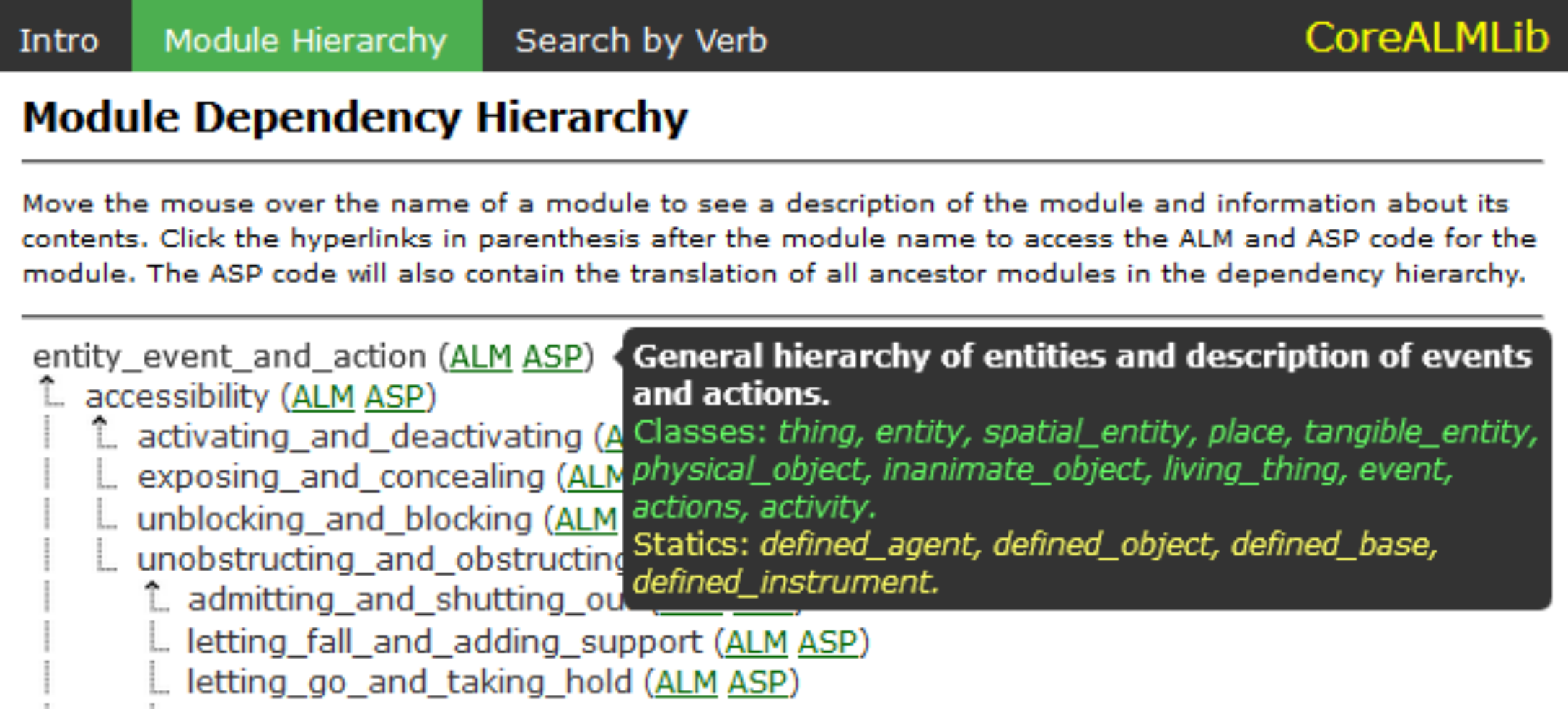}}
\caption{Online Tool: Viewing the Module Dependency Hierarchy}
\label{module_hierarchy}
\end{figure}

The \corealm\ library is available online at the web page http://tinyurl.com/z6n9fmx. 
Users can download the entire library and a prototype translator from \alm\ into
ASP from the main page. An additional page allows users to 
see the module dependency hierarchy and download individual modules (or their translation into ASP).
When the user moves the mouse over the title of a module, a description of the module
is displayed, containing the general purpose of the module, together with the list of 
action classes and functions declared in it (those that are new compared to ancestor modules in the 
hierarchy). An example can be seen in Figure \ref{module_hierarchy}.

\begin{figure}[!htbp]
\centering
\setlength\fboxsep{0pt}
\setlength\fboxrule{0.5pt}
\fbox{\includegraphics[width=1\textwidth]
{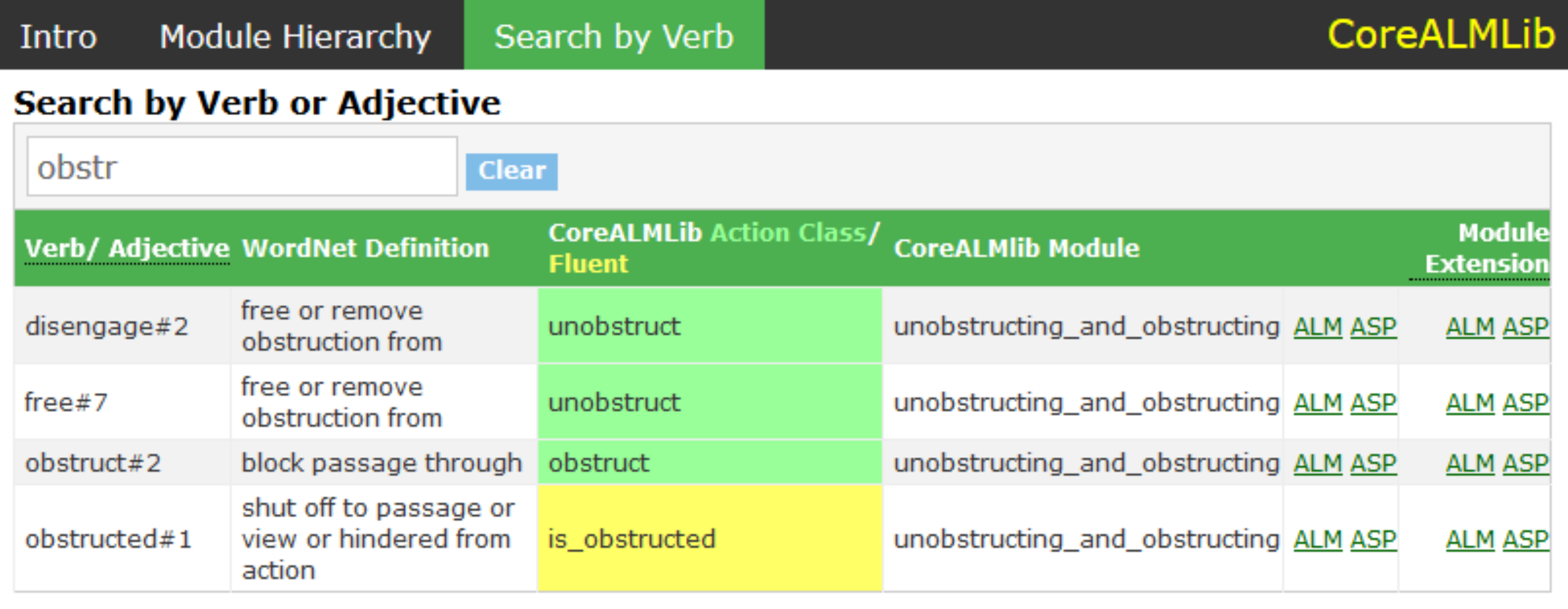}}
\caption{Online Tool: Searching by English Verb or Adjective}
\label{search}
\end{figure}

A third page contains a lookup table for which the keys are English verbs or adjectives 
accompanied by WordNet sense numbers -- see Figure \ref{search}. These words were extracted from the 
information about WordNet 2.0 synonym sets of \aclib\ action and state classes.
Additionally, the table contains the definition of the word sense from WordNet,
the name of the corresponding \corealm\ action class or fluent and the module in which it can be found
with links to the \alm\ and ASP code,
and links to an extended module with optional axioms, if it exists.
The table is searchable by verb/ adjective and WordNet definition (i.e., the two leftmost 
columns), as it can be seen in the example
in Figure \ref{search}.

\subsection{Testing and Using \corealm}

We tested the library using test cases provided in \clib.
Additionally, we encoded some scenarios that required the use of
multiple modules for representation.

In order to create an \alm\ encoding of a particular domain,
the user would first determine which \corealm\ modules are relevant.
For that, the ``Search by Verb or Adjective'' capability of the online tool should be used.
Afterward, the user can either use \alm\ or ASP to expand
the general knowledge from the library modules with knowledge particular to the domain.
We explain here the first option and take as an example the following text:

``{\em The wrestler restrained his opponent.}'' 

\noindent
By searching for the verb ``restrain'' in the online tool, the user would determine that the 
relevant \corealm\ action class is $restrain$ and that it can be found in the 
module $unrestraining\_and\_$ $restraining$. She would create a system description
titled $wrestler\_and\_opponent$ and import into its theory this library module:

$
\begin{array}{l}
{\bf system \ description} \ wrestler\_and\_opponent\\
\ \ {\bf theory} \ wrestler\_and\_opponent\\
\ \ \ \ {\bf import\ from }\ coreALMlib \ {\bf module} \ unrestraining\_and\_restraining
\end{array}
$

\noindent
The import statement is equivalent to copying the contents of the imported module and 
of all its ancestors into the theory (e.g., $unobstructing\_and\_obstructing$).
Next, the user would create a structure defining the 
entities mentioned in the sentence as instances of classes in the theory

$
\begin{array}{l}
\ \ {\bf structure} \ wrestler\_and\_opponent\\
\ \ \ \ {\bf instances}\\
\ \ \ \ \ \ wrestler, opponent \ {\bf in} \ living\_entity
\end{array}
$

\noindent
and the event in the sentence as an instance of $restrain$ in which the agent (the ``doer'') 
is the $wrestler$ and
the object (the one affected by the action) is the $opponent$.

$
\begin{array}{l}
\ \ \ \ \ \ r \ {\bf in} \ restrain\\
\ \ \ \ \ \ \ \ agent(wrestler) = true\\
\ \ \ \ \ \ \ \ object(opponent) = true
\end{array}
$

\noindent
(Note that even this simple domain cannot be encoded using the MAD libraries 
in Erdo\v{g}an's thesis \citeyear{thesiser08} in a comparably simple way.)
The system description would be translated 
using the prototype translator from \alm\ into ASP.
The user would have to add to the resulting logic program 
a predefined module for temporal projection and a history \cite{gk14}.
For our scenario, the history would specify that 
action $r$ was observed to have happened at time step 0:

$
\begin{array}{l}
\ \ \ \ \ \ hpd(r, 0)
\end{array}
$

\noindent
and that neither the wrestler nor the opponent were initially restrained:

$
\begin{array}{l}
\ \ \ \ \ \ obs(is\_restrained(wrestler), false, 0)\\
\ \ \ \ \ \ obs(is\_restrained(opponent), false, 0)\mbox{.}
\end{array}
$

\noindent
Answer sets of the program will indicate that, as a result of action $r$
happening at time step 0, the opponent will be restrained at the end of the story.

If we added the question
``{\em What would need to happen in order for the opponent to be able to move freely?}''
then the user would need to expand the theory of the initial system description
by at least importing the \corealm\ module $motion$ that contains the description
of action $move$ mentioned in the question. Note that this module specifies that
the action cannot be executed if the object is restrained.
This is in fact a planning problem, and the solution will depend on the action instances
added to the structure. The user would expand the initial structure by at least:

$
\begin{array}{l}
\ \ \ \ \ \ m \ {\bf in} \ move \\ 
\ \ \ \ \ \ \ \ object(opponent) = true
\end{array}
$

\noindent
and

$
\begin{array}{l}
\ \ \ \ \ \ u(X, Y)  \ {\bf in} \ unrestrain\\
\ \ \ \ \ \ \ \ agent(X) = true\\
\ \ \ \ \ \ \ \ object(Y) = true
\end{array}
$

\noindent
After adding a planning module \cite{gk14} to the ASP translation of this extended system description,
two possible solutions would be found: either that the 
wrestler unrestrains his opponent (i.e., execute $u(wrestler, opponent)$)
or that the opponent unrestrains himself (i.e., execute $u(opponent, opponent)$).
Note that the KM inference engine cannot perform planning.


\section{Conclusions and Future Work}
\label{conclusions}

In this paper we have described a library of core commonsense knowledge about dynamic domains
in modular action language \alm. The library, called \corealm, is obtained by translating
a big part of an established upper ontology, \clib, into \alm. We have provided 
a translation from the language of \clib\ into \alm, 
thus narrowing the gap between different KR languages. We have discussed
the methodology used to group resulting declarations and axioms into \alm\ modules. 
We have created an online tool that allows users to view the module dependency hierarchy
and search for relevant modules based on English verbs or adjectives. Finally, we have
discussed how the library can be used in practice to reason about dynamic domains. 

We believe that \corealm\ is, at least in part, more elaboration tolerant than its \clib\ counterpart,
specifically with respect to the description of action effects and preconditions.
On the other hand, restrictions on the values of attributes of a 
class are more elegantly expressed in \clib. This indicates possible syntactic additions to \alm.
\corealm\ can be seamlessly coupled with reasoning algorithms in ASP to solve
complex tasks such as diagnosis \cite{bg03} and planning, that cannot be answered in the inference engine of \clib.

\corealm\ can also further motivate the research on libraries in \alm,
including the structuring of knowledge; 
further means for finding relevant modules; 
and providing to the user information about the contents of a module. 
It can also drive the investigation on how to allow users to select only
those pieces of a module (declarations or axioms) that are relevant to
a specific dynamic domain, which we believe to be an interesting research question.


\medskip
\noindent
\textbf{Acknowledgments.} This research was funded in part by SRI International and Vulcan Inc.
The author is also grateful to Michael Gelfond for his feedback.

\bibliographystyle{acmtrans}
\bibliography{alm-and-clib}

\end{document}